\documentclass[runningheads]{llncs}
\usepackage{graphicx,multirow,tabulary}
\usepackage{comment}
\usepackage{amsmath,amssymb} 
\usepackage{color}
\usepackage{booktabs} 
\usepackage{multirow} 
\def\x{$\times$}
\usepackage{bm}
\usepackage[caption=false]{subfig}

\begin{document}
\pagestyle{headings}
\mainmatter

\title{Local Grid Rendering Networks for \\ 3D Object Detection in Point Clouds} 

\titlerunning{Local Grid Rendering Networks}
\author{Jianan Li, Jiashi Feng}
\authorrunning{J. Li et al.}
\institute{National University of Singapore \\
\email{lijianan15@gmail.com, elefjia@nus.edu.sg}}
\maketitle

\begin{abstract}
The performance of 3D object detection models over point clouds highly depends on their capability of modeling local geometric patterns. Conventional point-based models exploit local patterns through a symmetric function (e.g. max pooling) or based on graphs, which easily leads to loss of fine-grained geometric structures. Regarding capturing spatial patterns, CNNs are powerful but it would be computationally costly to directly apply convolutions on point data after voxelizing the entire point clouds to a dense regular 3D grid. In this work, we aim to improve performance of point-based models by enhancing their pattern learning ability through leveraging CNNs while preserving computational efficiency. We propose a novel and principled Local Grid Rendering (LGR) operation to render the small neighborhood of a subset of input points into a low-resolution 3D grid independently, which allows small-size CNNs to accurately model local patterns and avoids convolutions over a dense grid to save computation cost. With the LGR operation, we introduce a new generic backbone called LGR-Net for point cloud feature extraction with simple design and high efficiency. We validate LGR-Net for 3D object detection on the challenging ScanNet and SUN RGB-D datasets. It advances state-of-the-art results significantly by \textbf{5.5} and \textbf{4.5} mAP, respectively, with only slight increased computation overhead. 
 
\end{abstract}

\section{Introduction}
Detecting objects in point clouds obtained from rapidly developing 3D scanners is an important first step for 3D scene understanding, and benefits various real-world applications such as autonomous navigation~\cite{gomez2016pl}, housekeeping robots~\cite{oh2002development}, and augmented/virtual reality~\cite{park2008multiple}.
However, unlike RGB images, point cloud data have own inherent properties --- they are sparse, irregular, with non-uniform density and lack of visual appearance information, which cause difficulty for object localization and recognition.
How to effectively extract local patterns in point data to assist detection still remains challenging.

\begin{figure}[t]
\centering
\includegraphics[width=\linewidth]{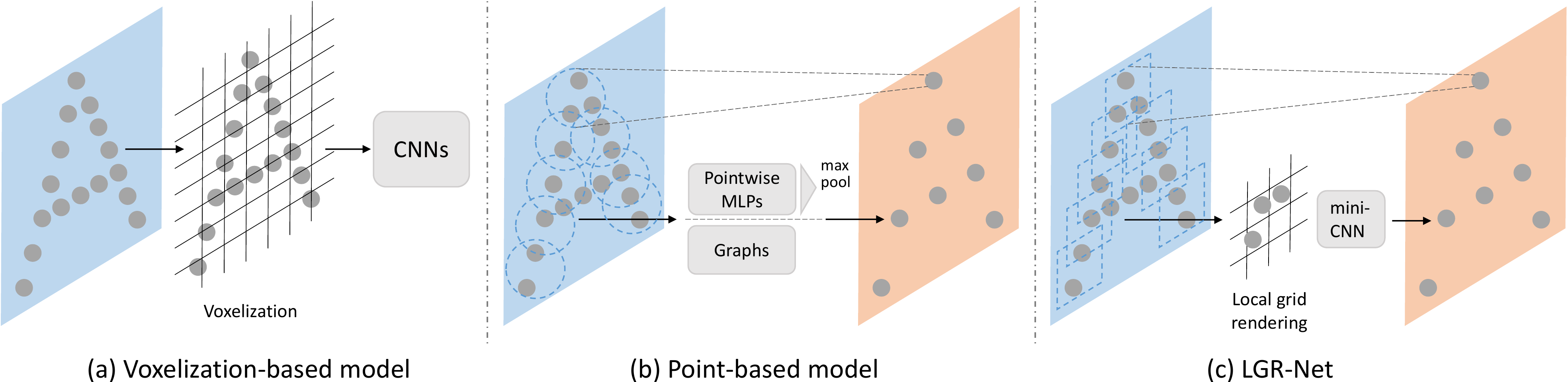}
\caption{\textbf{Approaches for learning geometric patterns in point clouds.} (a) Voxelization-based model. The entire input points are directly voxelized into a dense grid. Geometric patterns are then learned through CNNs, which is accurate but computationally expensive.
(b) Point-based model. A subset of input points is sampled and enriched with local context through a symmetric function (e.g. max pooling) or based on graphs, which is fast but tends to miss fine-grained geometry.
(c) Our LGR-Net. It benefits from both (a) and (b) by sampling local point sets and converting each of them to a small grid independently, 
which allows using powerful small-size CNNs to effectively capture fine-grained geometry yet preserves computational efficiency by only convolving on small grids derived from sensed regions
}
\label{fig:motivation}
\vspace{-0.2in}
\end{figure}

Many approaches~\cite{zhou2018voxelnet,song2016deep,hou20193d,qi2017pointnet++,wang2019dynamic,qi2019deep} have been developed for learning local geometric structures in the point clouds, which is crucial to detecting objects.
One line of works~\cite{zhou2018voxelnet,le2018pointgrid,song2016deep,hou20193d} voxelize the irregular point clouds to a regular 3D grid and apply 3D CNNs to extract features from neighboring voxels progressively, as shown in Fig.~\ref{fig:motivation}(a). 
The advantage of such solutions lies in utilizing the prominent power of convolutional kernels for learning spatial patterns to detect geometric structures. 
However, these approaches require to transform the entire point clouds to volumetric representation, where the sparsity in point data does not get exploited, thus suffer high computational cost from applying 3D convolutions over dense grids mainly occupied with empty voxels.

On the other hand, some works~\cite{qi2017pointnet++,wang2019dynamic,lan2019modeling,qi2019deep} accept raw points without voxelization, as illustrated in Fig.~\ref{fig:motivation}(b). 
Such point-based models abstract local patterns from small neighborhoods of subsampled points through either point-wise Multi-layer Perceptrons (MLPs) followed by a symmetric function (e.g. max pooling~\cite{qi2017pointnet++}), 
or based on graphs with MLPs applied on each graph edge~\cite{wang2019dynamic,lan2019modeling}.
In contrast to voxelization-based models, point-based models respect the inherent sparsity of point data and enable efficient computation by only computing on sensed regions.
However, using discrete MLPs leads to loss of relations among many points, thus tends to miss the fine-grained geometry.

In this work, we aim to improve performance of point-based models by strengthening their pattern learning ability with 3D CNNs. 
The key challenge is that point-based models directly operate on sparse and irregular points and hence are not compatible with standard convolutions. 
Our key idea is to develop a new local reshaping approach of point clouds such that they are compatible with regular convolutions without incurring too much computation overhead. 
To this end, we operate on subsampled points from the input to leverage its sparsity, and introduce a novel and principled Local Grid Rendering (LGR) operation to render a small neighborhood of each sampled point to a regular grid of low-resolution independently. In this way, a small-size CNN is allowed to only compute on a set of small grids from sensed regions to abstract fine-grained patterns efficiently. See the illustration in Fig.~\ref{fig:motivation}(c).

Concretely, we implement the above idea with three steps.
1) \emph{Data structuring}. A subset of input points is sampled as centroids of sensed regions, and neighboring points  to each region centroid are queried to construct local point sets. 
2) \emph{LGR operation}. Each local point set is converted to a small regular grid independently through an interpolation function. 
3) \emph{Perceptual feature extraction}. With the LGR operation, an efficient mini-CNN perceives and abstracts spatial patterns in each rendered grid into a feature vector.

We wrap these separate steps into an integrated Set Perception (SP) module, which abstracts a set of input points to produce a new set with fewer points, and enriches them with local context perceived by CNNs in a certain neighborhood.
By applying the SP module repeatedly, fewer points with context from larger neighboring regions can be obtained progressively. 
With the newly designed SP module, we build a simple yet efficient backbone network, named \emph{LGR-Net}, for effective feature extraction for point clouds.
Our LGR-Net is superior in learning geometric structures by applying the LGR operation, which makes irregular point data compatible with 3D CNNs locally. 
Meanwhile, it also enables efficient computation in that only sensed regions are projected to small grids, which avoids redundant convolutions in empty space.

Our proposed LGR-Net is generic and applicable to various point-based models. 
In this work, we apply LGR-Net to build a 3D detection model on top of the recent successful deep Hough voting framework VoteNet~\cite{qi2019deep}.
We perform evaluations on two challenging 3D indoor object detection datasets,  ScanNet~\cite{dai2017scannet} and SUN RGB-D~\cite{song2015sun}.
Our model achieves state-of-the-art results on both of them with significant improvements ($+\bm{5.5}$ and $+\bm{4.5}$ mAP, respectively) over the prior VoteNet, bringing only slightly increased computation overhead.
It is evidenced that incorporating CNNs through LGR enables more effective abstraction of local patterns than conventional point-based models, thus benefits the detection of 3D objects in point clouds.

In summary, this work makes the following contributions:
\begin{itemize}
    \item We propose a novel LGR operation that projects local point sets to small 3D grids independently. It allows using 3D CNNs to abstract local geometric patterns explicitly and efficiently.
    \item Based on the LGR, we introduce a new backbone network (LGR-Net) to the community, which is simple and efficient.
    \item Our model establishes new state-of-the-art on the ScanNet and SUN RGB-D datasets.
\end{itemize}

\section{Related Work}

\subsubsection{Point-based Models for Point Clouds}
Modeling geometric relations among points is a fundamental step for analyzing point clouds.
Some earlier works~\cite{bronstein2010scale,golovinskiy2009shape,gomes2013efficient} develop hand-crafted feature descriptors to capture geometric patterns.
Recently, point-based models are proposed to learn deep features on raw point data. 
PointNet~\cite{qi2017pointnet} and PointNet++~\cite{qi2017pointnet++} use point-wise MLPs followed by max pooling to aggregate point features. 
Some other works~\cite{wang2019dynamic,lan2019modeling} construct a local neighborhood graph based on the neighbors of each point, and perform MLPs on the edges to learn their relations.
Though enjoying a high processing speed, these point-based models endure implicit abstraction of local geometry.
Considering explicit modeling of geometric structures, InterpConv~\cite{mao2019interpolated} directly convolves on irregular point clouds by interpolating point features to the neighboring weights of discrete convolutional kernels.
Our approach is different from InterpConv in that we first convert the point set in a local region into a small regular grid, which allows using a mini-CNN flexibly with arbitrary numbers and kernel sizes of successive convolutional layers to fully exploit local patterns.

\vspace{-0.05in}
\subsubsection{Point-based Models in 3D Object Detection}
Recent development of real-time applications such as autonomous driving and robotics has motivated increasing attention to 3D object detection in point clouds.
Some early works~\cite{Indoor2012,litany2015asist,song2014sliding,avetisyan2019scan2cad} rely on template matching to localize objects by aligning a set of CAD models to 3D scans. 
MV3D~\cite{chen2017multi} projects 3D data to the bird’s eye view representation to generate candidate boxes.
With the demand for high processing speed, point-based models suited for raw point data have been working well for 3D object detection~\cite{lang2019pointpillars,shi2019pointrcnn,qi2019deep} and also semantic and instance segmentation~\cite{yi2019gspn,choy20194d,graham20183d}.
Among these models, PointRCNN~\cite{shi2019pointrcnn} develops a two-stage detector which generates 3D proposals and refines them directly from input points.
Notably, VoteNet~\cite{qi2019deep} constructs an end-to-end detection framework by incorporating PointNet++~\cite{qi2017pointnet++} and Hough voting process, hitting new state-of-the-art when applied to indoor scenes~\cite{dai2017scannet,song2015sun} with just point data input. 

\vspace{-0.05in}
\subsubsection{Voxelization-based Models for Point Clouds}
Voxelization-based models~\cite{zhou2018voxelnet,qi2016volumetric,wu20153d,engelcke2017vote3deep} convert input point clouds into a regular 3D grid upon which 3D CNNs are utilized for feature extraction. 
The limitation of such models is that a sparse grid suffers quantization artifacts while a dense grid leads to exponentially growing computational complexity. 
Indexing techniques have been used to operate on non-uniform grids~\cite{klokov2017escape,riegler2017octnet}, but they focus more on subdivisions of point clouds than on modeling geometric structures. 
Voxelization-based models have also been applied to 3D object detection.
VoxelNet~\cite{zhou2018voxelnet}, DSS~\cite{song2016deep} and 3D-SIS~\cite{hou20193d} divide input point clouds into equally spaced 3D grids for unified feature extraction and bounding box prediction using 3D CNNs.
However, such methods are computationally expensive due to redundant convolutions over dense grids occupied with many empty voxels.

\section{Method}
We propose a simple yet efficient backbone LGR-Net for point cloud feature extraction based on a novel Local Grid Rending (LGR) operation, and apply LGR-Net to 3D object detection. In following, we will first elaborate on our LGR operation and then LGR-Net, and finally explain its application to 3D objection detection.

\begin{figure}[t]
\centering
\includegraphics[width=\linewidth]{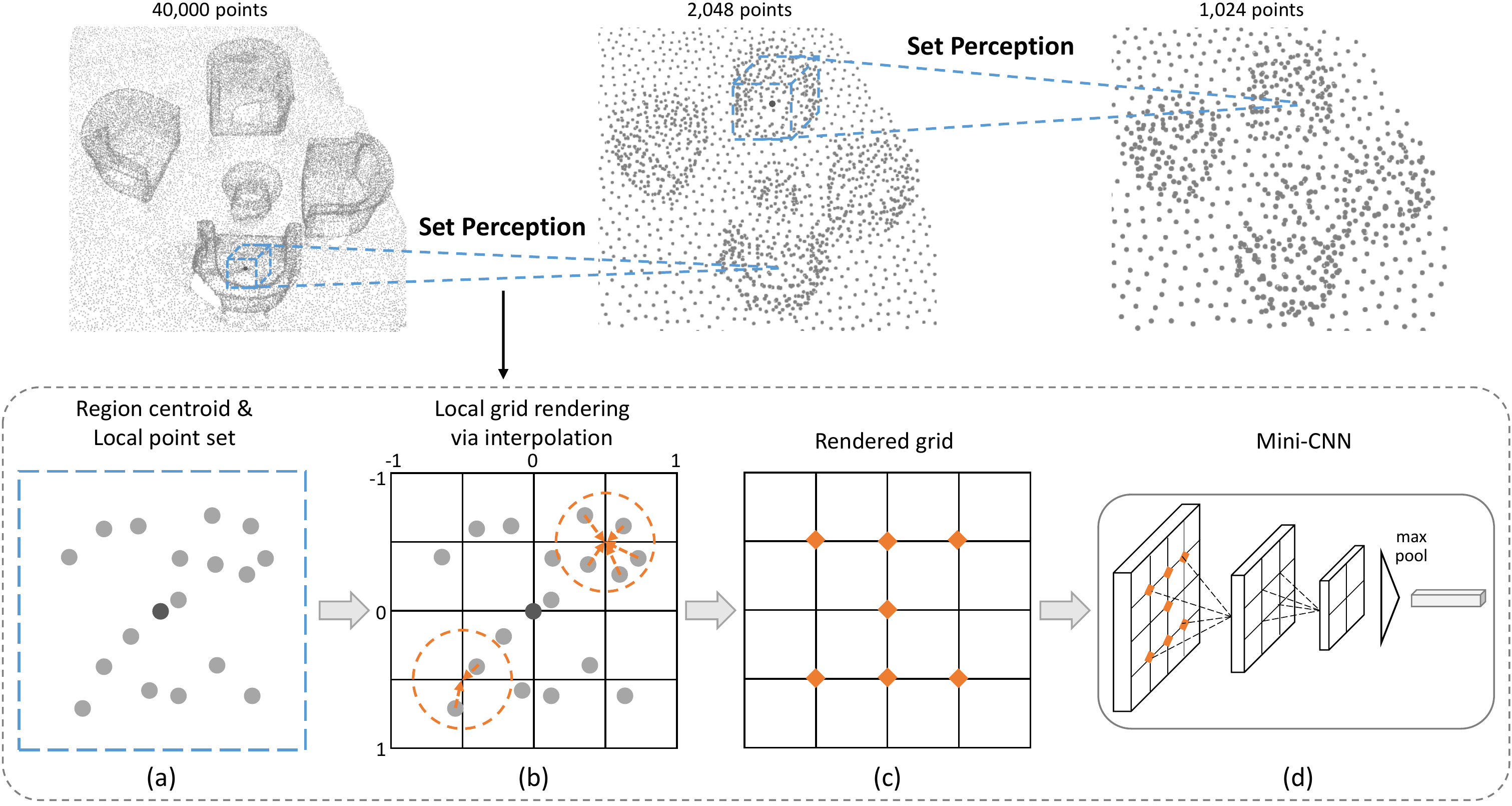}
\caption{\textbf{Feature abstraction from point clouds with LGR.} Input points are abstracted through a hierarchy of set perception processes, each of which enriches a subsampled set of input points
with local context.
Staged procedures for set perception are presented in 2D for clearness. 
(a) Sampling a subset of input points as region centroids and querying their neighboring points to construct local point sets. 
(b) Projecting each local point set to a small grid independently. 
The activation on each voxel is computed as a weighted average of features of its neighboring points within a preset radius to deal with non-uniform point density. 
(c) The Rendered grid. Voxels with a non-zero response are highlighted. 
(d) Applying a mini-CNN on the rendered grid to perceive and abstract local patterns into a feature vector}
\label{fig:grid_rendering}
\vspace{-0.1in}
\end{figure}

\subsection{Set Perception with LGR}
We progressively abstract local patterns from input points with three steps including \emph{data structuring}, \emph{Local Grid Rendering (LGR)} and \emph{perceptual feature extraction}, which are integrated into a Set Perception (SP) module.  
See the illustration in Fig.~\ref{fig:grid_rendering}.
The SP module perceives and abstracts a set of input points to produce a new set with fewer elements. 
It takes as input $N$ points, each of which with xyz-coordinates and $C$-dim features, forming input data size $N$\x$(3+C)$. 
It outputs $N'$ subsampled points of size $N'$\x$(3+C')$ comprising xyz-coordinates and new $C'$-dim features enriched with local context. The three steps are explained one by one at below.

\subsubsection{Data Structuring}
Given input $N$ points of size $N$\x$(3+C)$, we sample a subset of $N'$ points as centroids of sensed regions. 
We use Farthest Point Sampling (FPS)~\cite{qi2017pointnet++} to get a better coverage of the entire input points compared to random sampling. 
We then select $K$ neighboring points for each region centroid, forming $N'$ groups of local point sets. Each point set represents geometric structures corresponding to a local region. 
Output data size is $N'$\x$K$\x$(3+C)$.

Concretely, we adopt Cube query~\cite{xu2019grid} to construct a cube region with a preset half-edge length $E$ centered at each region centroid, and randomly pick $K$ points within the cube to form a local point set. 
We translate the coordinates of $K$ points in each set into a local frame relative to its region centroid, by firstly subtracting the centroid's coordinates and then divided by the half-edge length $E$, so that the translated point coordinates $\bm{\theta}$ fall in the interval $[-1,1]$.

\subsubsection{Local Grid Rendering}
Suppose the $K$ points in a local point set are parameterized as $\left \{(\bm{f_{1}}, \bm{\theta_{1}}), ..., (\bm{f_{K}}, \bm{\theta_{K}})\right \}$, in which the feature vector $\bm{f_i} \in \mathbb{R}^{C}$ and coordinates $\bm{\theta_i} = (x_i, y_i, z_i) \in [-1,1]^{3}$ for the point $i$.
The novel LGR operation rasterizes each local point set into a regular 3D grid independently through an interpolation function. 
Since each grid only represents geometric structures of a local neighborhood, the grid resolution is allowed to be very small.
The rasterization is performed as follows. 

Each point in the set can be rendered into its own grayscale 3D grid $\bm F_{\bm{\theta}} (x,y,z)$ of preset resolution $W$\x$H$\x$L$ (width\x height\x length in voxels), where voxel coordinates are uniformly spaced in the interval $[-1,1]$ so as to be compatible with the coordinate range of the point.

For a single point $i$, we implement an interpolation kernel $k$ for its rasterization. 
Its interpolated response on voxel $\bm{v}=(x, y, z)$ in the grayscale gird is computed as
\begin{equation}
    \bm{F}_{\bm{\theta_i}}(x,y,z) = k\left(  d(\bm{\theta_i}, \bm{v})  \right),
    \label{eqn:grid_radius}
\end{equation}
where $d(\cdot)$ computes the distance between the voxel and the point $i$. 
We use Euclidean distance in our experiments. 
The interpolation kernel $k(\cdot)$ is defined as 
\begin{equation}
    k(m) = \max\left(0, 1-(m/r)^p\right),
    \label{eqn:grid_kernel}
\end{equation}
where $r$ is a preset radius within which the voxels can get a non-zero interpolated response from the point $i$. 
We set $r$ as half the diagonal of a voxel cell (discussed in Sec.~\ref{sec:exp_aggregation}). 
The parameter $p>0$ adjusts the rate at which interpolated response is decreased as the distance between the voxel and the point $i$ increases (in default we use $p=1$).

The rendering output is a multi-channel 3D grid $\bm I$ of dimension $W$\x$H$\x$L$\x$C$, where each channel corresponds to one of the $C$ channels of point features.
Voxel $(x,y,z)$ of $\bm I$ is a feature activation vector for that voxel, and is computed as
\begin{equation}
    \bm I(x,y,z,c) =\frac{\sum_{i=1}^{K}{\bm F_{\theta_i}(x,y,z)}f_{i,c}}{\sum_{i=1}^{K} \bm F_{\theta_i}(x,y,z)},
    \label{eqn:grid_interpolate}
\end{equation}
where $f_{i,c}$ is the $c$-th channel feature of the point $i$. As a result, though points in a local point set are distributed non-uniformly across space, the given weighted average aggregation of point features ensures that the scale of activation on different voxels in the grid is invariant to varying point density.

\subsubsection{Perceptual Feature Extraction}
This step perceives and abstracts spatial patterns in a rendered grid $\bm I$ into a $C'$-dim feature vector by performing a function $f: \bm I \rightarrow \mathbb{R}^{C'}$:
\begin{equation}
    f(\bm I) = g\left(h(\bm I)\right),
    \label{eq:pointnet}
\end{equation}
where $h(\cdot)$ and $g(\cdot)$ are functions for feature embedding and spatial pooling respectively.

Concretely, we apply two 3D convolutional layers followed by a global max pooling layer, forming a shared mini-CNN, to abstract patterns in each grid. 
Since the mini-CNN only consists of a handful of convolutional layers, which are performed on a set of low-resolution grids, massive convolutions over dense grids are thus avoided to remain efficiency.

\subsection{LGR-Net Architecture}
The SP module samples a subset of $N'$ points from the input points and enriches each sampled point with local context from a neighboring cube region with a preset half-edge length $E$. 
By stacking SP modules with decreasing $N'$ and increasing $E$, fewer points with context from larger regions can be obtained progressively.
We thus build a new backnone LGR-Net by forming a hierarchy of SP modules for point feature extraction.

Concretely, LGR-Net comprises four SP modules with decreasing $N'$ including $2048$, $1024$, $512$, $256$, and increasing $E$ including $0.15$, $0.3$, $0.6$, $1.0$ in meters respectively.
In addition, two Feature Propagation (FP) layers~\cite{qi2017pointnet++} upsample the output from the 4th SP module back to $1024$ points with $256$-dim features, by interpolating the features on input points to output points (each output point’s feature is the inverse distance weighted average of its three nearest input points’ features). 
Details of the LGR-Net architecture are specified in Table~\ref{table:network_arc}.

\newcommand{\blocka}[2]{\multirow{2}{*}{\(\left[\begin{array}{l}\text{3$\times$3$\times$3, #1}\\[-.1em] \text{3$\times$3$\times$3, #2} \end{array}\right]\)}}
\begin{table}
\setlength{\tabcolsep}{5.0pt}
\renewcommand{\arraystretch}{1.3}
\begin{center}
\caption{\textbf{Details of LGR-Net architecture.} Feature propagation layers are not shown here. Module specifications (1st column) are given, including input data size, (number $N'$ of subsampled points, half-edge length $E$ for cube query, number $K$ of queried neighboring points), kernel size and channel number of 3D convolutional layers, and output data size}
\label{table:network_arc}
\begin{tabular}{c|c|c|c|c}
    \toprule[0.8pt]
	Module & SP$_1$ & SP$_2$ & SP$_3$ & SP$_4$ \\ 
    \midrule[0.8pt] 
	Input size  & N\x (3+1) & 2048\x 128 & 1024\x 256 & 512\x 256  \\\hline
	 Sampling  & (2048, 0.15, 64)  & (1024, 0.3, 32)  & (512, 0.6, 16)  & (256, 1.0, 16) \\\hline
	\multirow{3}{*}{Mini-CNN}  & \blocka{64}{128} & \blocka{128}{256} & \blocka{128}{256} & \blocka{128}{256} \\
	 &  &  &  & \\
	 & max pool  & max pool  & max pool  & max pool \\\hline
     Output size & 2048\x 128 & 1024\x 256 & 512\x 256  & 256\x 256  \\
	\bottomrule[0.8pt]
\end{tabular}
\end{center}
\vspace{-0.2in}
\end{table}

\subsection{Application for Detection}
Our LGR-Net is generic. We apply it to object detection considering its ability to extract fine-grained local geometric structures that are crucial to detection performance.
We build a point-based detection model based on the recent successful VoteNet~\cite{qi2019deep} framework, which directly works on raw point clouds and outputs object proposals in one forward pass by using a backbone network, a voting module and a proposal module.

Concretely, we apply LGR-Net as the backnone network to output a subset of input points featured by local patterns, which are considered as seed points. 
The voting module~\cite{qi2019deep}, implemented as a MLP, generates votes from each seed independently.
Every vote is a 3D point with its coordinates regressed to approach the object center, and also a refined feature vector.
The proposal module, implemented as a set abstraction layer~\cite{qi2017pointnet++}, groups the votes into clusters and aggregates their features to generate object proposals. 
These proposals are further classified and NMSed to output final 3D bounding boxes.

\section{Experiments}
In this section, we firstly compare our method with previous state-of-the-arts on two popular 3D object detection benchmarks of indoor scenes. We then provide experimental analysis to validate our design choices. Finally, qualitative results along with discussions are given.

\subsection{Implementation Details}
Our detection model is based on the VoteNet~\cite{qi2019deep} framework, thus we adopt the same input and data augmentation scheme as in VoteNet for a fair comparison.
Specifically, the network input is a set of $N$ points randomly subsampled from a popped-up depth image ($N$=$20k$) or a 3D scan (mesh vertices, $N$=$40k$) on-the-fly.
Each point is represented by a 4-dim vector of xyz-coordinates with an additional height feature indicating the distance to the floor~\cite{qi2019deep}. 
Several augmentations are applied to the points, including random flipping in both horizontal directions, random rotation by $\text{Uniform}[-5^{\circ},5^{\circ}]$ around the upright-axis, and random scaling by $\text{Uniform}[0.9,1.1]$.

The entire network is trained end-to-end from scratch. 
We adopt Adam optimizer with batch size $8$ and an initial learning rate of $0.001$ which is decreased by $10\times$ after $80$ epochs and by another $10\times$ after $120$ epochs. 
For inference, we perform 3D non-maximum suppression on the generated object proposals with an IoU threshold of $0.25$. 
The evaluation protocol is Average Precision (AP) following Song et al.~\cite{song2016deep}.

\begin{table}[t]
\fontsize{8}{8}\selectfont
\setlength{\tabcolsep}{0.9pt}
\begin{center}
\caption{\textbf{3D object detection results on SUN RGB-D val set.} Evaluation is measured by average precision with 3D IoU threshold of 0.25~\cite{song2015sun}, and 
conducted on SUN RGB-D V1 data for fair comparisons with previous methods}
\label{table:sunrgbd}
\begin{tabular}{l|c|cccccccccc|c}
\toprule
Methods & Input & bathtub & bed & bkshf & chair & desk & dresser & ntstd & sofa & table & toilet & mAP \\ 
\midrule
DSS~\cite{song2016deep} & Geo+RGB & 44.2 & 78.8 & 11.9 & 61.2 & 20.5 & 6.4 & 15.4 & 53.5 & 50.3 & 78.9 & 42.1    \\
COG~\cite{ren2016three} & Geo+RGB & 58.3 & 63.7 & 31.8 & 62.2 & \textbf{45.2} & 15.5 & 27.4 & 51.0 & 51.3 & 70.1 & 47.6 \\
2D-driven~\cite{lahoud20172d} & Geo+RGB & 43.5 & 64.5 & 31.4 & 48.3 & 27.9 & 25.9 & 41.9 & 50.4 & 37.0 & 80.4 & 45.1  \\
PointFusion~\cite{xu2018pointfusion} & Geo+RGB & 37.3 & 68.6 & 37.7 & 55.1 & 17.2 & 23.9 & 32.3 & 53.8 & 31.0 & 83.8 & 45.4 \\ 
F-PointNet~\cite{qi2018frustum} & Geo+RGB & 43.3 & 81.1 & 33.3 & 64.2 & 24.7 & 32.0 & 58.1 & 61.1 & 51.1 & \textbf{90.9} & 54.0 \\
\midrule
VoteNet~\cite{qi2019deep} & Geo only & 74.4 & 83.0 & 28.8 & 75.3 & 22.0 & 29.8 & 62.2 & 64.0 & 47.3 & 90.1 & 57.7 \\ 
LGR-Net (ours)   & Geo only & \textbf{80.2} & \textbf{85.1} & \textbf{39.7} & \textbf{76.7} & 29.4 & \textbf{35.1} & \textbf{66.5} & \textbf{67.6} & \textbf{52.7} & 89.1 & \textbf{62.2} \\
\bottomrule
\end{tabular}
\end{center}
\vspace{-0.05in}
\end{table}

\begin{table}
\setlength{\tabcolsep}{4.0pt}
\begin{center}
\caption{ \textbf{3D object detection results on ScanNetV2 val set.}
Evaluation metric is mean average precision with 3D IoU threshold of 0.25 and 0.5}
\label{table:scannet}
\begin{tabular}{l|c|cc}
\toprule
Methods & Input & mAP@0.25 & mAP@0.5 \\ \hline    
DSS~\cite{song2016deep,hou20193d} & Geo + RGB & 15.2 & 6.8  \\
MRCNN 2D-3D~\cite{he2017mask,hou20193d} & Geo + RGB & 17.3 & 10.5 \\
F-PointNet~\cite{qi2018frustum,hou20193d} & Geo + RGB & 19.8 & 10.8 \\
GSPN~\cite{yi2019gspn} & Geo + RGB & 30.6 & 17.7 \\ 
\midrule
3D-SIS \cite{hou20193d} & Geo + 1 view & 35.1 & 18.7 \\ 
3D-SIS \cite{hou20193d} & Geo + 3 views & 36.6 & 19.0 \\
3D-SIS \cite{hou20193d} & Geo + 5 views & 40.2 & 22.5 \\ 
\midrule
3D-SIS \cite{hou20193d} & Geo only & 25.4 & 14.6 \\
VoteNet~\cite{qi2019deep} & Geo only & 58.6 & 33.5 \\
LGR-Net (ours) & Geo only & \textbf{64.1} & \textbf{42.0} \\ 
\bottomrule 
\end{tabular}
\end{center}
\vspace{-0.15in}
\end{table}

\subsection{Comparing with State-of-the-Arts}
\subsubsection{Benchmark Datasets.} 
\emph{SUN RGB-D}~\cite{song2015sun} contains $10,335$ single-view RGB-D images ($\sim$5,000 for training) with dense annotations of oriented 3D bounding boxes for $37$ object categories. 
We reconstruct point clouds from the depth images using the provided camera parameters, as in VoteNet. 
Following standard evaluation protocol~\cite{song2016deep}, we only train and report results on the $10$ most common categories.
\emph{ScanNetV2}~\cite{dai2017scannet} is a large-scale RGB-D dataset containing $1,513$ scans of over $707$ unique indoor scenes.
The scans are annotated with surface reconstructions, textured meshes and semantic and instance segmentation for $18$ object categories. 
We reconstruct input point clouds by sampling vertices from the reconstructed meshes and predict axis-aligned bounding boxes~\cite{qi2019deep}. 
We use $1,201$ scenes for training, and $312$ scenes for testing~\cite{dai2017scannet}.

\subsubsection{Methods in Comparison.}
2D-driven~\cite{lahoud20172d}, PointFusion~\cite{xu2018pointfusion} and F-PointNet~\cite{qi2018frustum} are 2D-driven 3D detection methods, which benefit from detection techniques for 2D images and use 2D information to reduce the search space in 3D.
Cloud of Gradients (COG)~\cite{ren2016three} designs new 3D HoG-like features to detect objects in a sliding shape manner.
MRCNN 2D-3D~\cite{he2017mask,hou20193d} estimates 3D bounding boxes by directly projecting instance segmentation results from Mask-RCNN~\cite{he2017mask} into 3D. 
GSPN~\cite{yi2019gspn} utilizes a generative model to generate object proposals for instance segmentation in point clouds. 

Notably, Deep Sliding Shapes (DSS)~\cite{song2016deep} and 3D-SIS~\cite{hou20193d} are voxelization-based detectors, which convert the entire input point clouds to dense grids and employ 3D CNNs to learn from both geometry and RGB cues for object proposal generation and classification. 
VoteNet~\cite{qi2019deep} is a point-based detector, which adopts PointNet++~\cite{qi2017pointnet++} as backbone to abstract and aggregate point features to generate proposals and classify them.

\subsubsection{Detection Performance.} 
Table~\ref{table:sunrgbd} and \ref{table:scannet} show detection results on SUN RGB-D and ScanNet respectively. 
Our model outperforms all prior arts by large margins on both benchmarks, i.e., at least \textbf{4.5} and \textbf{5.5} mAP increase, respectively.
Note that our model takes only point clouds as input, while most previous methods use both geometry and RGB or even multi-view RGB images.
Especially, Table~\ref{table:sunrgbd} shows our model achieves better results on nearly all categories, over both the best-performing point-based detector (VoteNet) and the voxelization-based detector (DSS).
Similar conclusions can be drawn from Table~\ref{table:scannet} that LGR-Net significantly outperforms both VoteNet and methods based on voxelization (DSS and 3D-SIS), evidencing the superiority of LGR-Net in local pattern abstraction and in turn detection.
In addition, Table~\ref{table:scannet} shows that the stricter criterion of mAP@0.5 is greatly boosted by \textbf{8.5}, suggesting our model advances object localization benefiting from the learned fine-grained geometry.

\subsubsection{Speed.}
Our model is computationally efficient since it only performs small-size CNNs on a set of grids of low-resolution, thanks to the LGR operation.
Table~\ref{tab:time} depicts that our model is more than \textbf{8}$\times$ times faster than voxelization-based 3D-SIS which applies CNNs on a dense grid derived from the entire point clouds.  
In addition, our model is comparable to the point-based VoteNet in terms of speed, while achieving significant improvement in detection performance.

\begin{table}
\begin{center}
\setlength{\tabcolsep}{10.0pt}
\caption{\textbf{Processing time per scan on ScanNetV2.} Our method is more than \textbf{8}$\times$ times faster than voxelization-based 3D-SIS~\cite{hou20193d} and comparable to point-based VoteNet~\cite{qi2019deep}}
\label{tab:time}
\begin{tabular}{l|c|c|c}
\toprule
    Methods & 3D-SIS~\cite{hou20193d} & VoteNet~\cite{qi2019deep} & LGR-Net (ours) \\ \midrule
    Time   & $2.85$s & $0.14$s & $0.33$s \\ \bottomrule
\end{tabular}
\end{center}
\vspace{-0.3in}
\end{table}

\subsection{Ablation Studies}
\subsubsection{How does LGR help?} 
We argue that leveraging 3D CNNs thanks to the LGR operation enables better abstraction of fine-grained structures than conventional point-based models.
However, directly analyzing learned features to support our argument is not trivial. 
Fortunately, the backbone network integrates learned local patterns into the features of a set of output points for predicting object proposals. 
Since fine-grained structures are crucial to correctly localizing and recognizing objects, an output point enriched with such features (called informative point here) is more likely to generate good votes, and in turn a good object proposal (with correct class and over $0.5$ IoU with ground-truth).
Therefore, a backbone better at capturing fine-grained structures should produce more informative points given a fixed number of output points.

\begin{figure}[t]
	\begin{center}
		\includegraphics[width=0.98\linewidth]{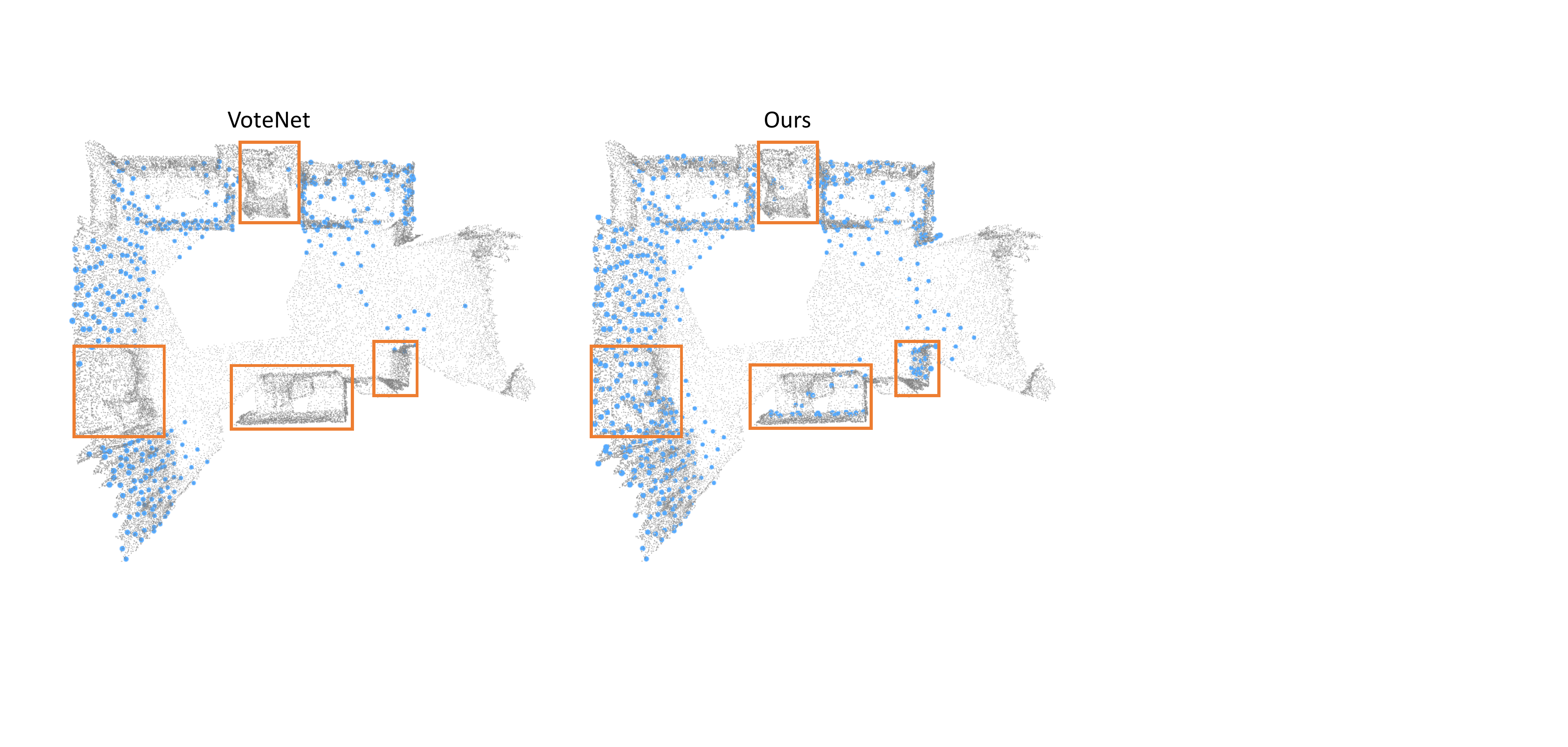}
	\end{center}
	\vspace{-0.2in}
	\caption{\textbf{LGR helps abstract fine-grained geometry.} We trace points (output by the backbone) that generate good votes which in turn generate good object proposals, and overlay such points (in blue) on top of input ScanNet scene points. Our model shows a broader coverage of the scene, especially on some small objects (in orange box) missed by VoteNet, proving LGR-Net better preserves fine-grained geometry and thus benefits detection}
	\label{fig:good_seeds}
	\vspace{-0.1in}
\end{figure}

Fig.~\ref{fig:good_seeds} demonstrates this phenomenon. 
We analyze points output by the backbone, and trace those points that can generate an accurate object proposal.
One can see that VoteNet (left) fails to produce informative points on some objects (especially small ones) to support detection.
In comparison, our model (right) offers more informative points with a denser coverage of the scene, especially on objects that VoteNet has missed, evidencing its superiority in exploiting fine-grained geometry.

\subsubsection{Effect of Feature Aggregation}
\label{sec:exp_aggregation}
\begin{figure}[t]
\begin{minipage}[c]{0.50\linewidth}
\begin{center}
    \includegraphics[width=0.96\linewidth]{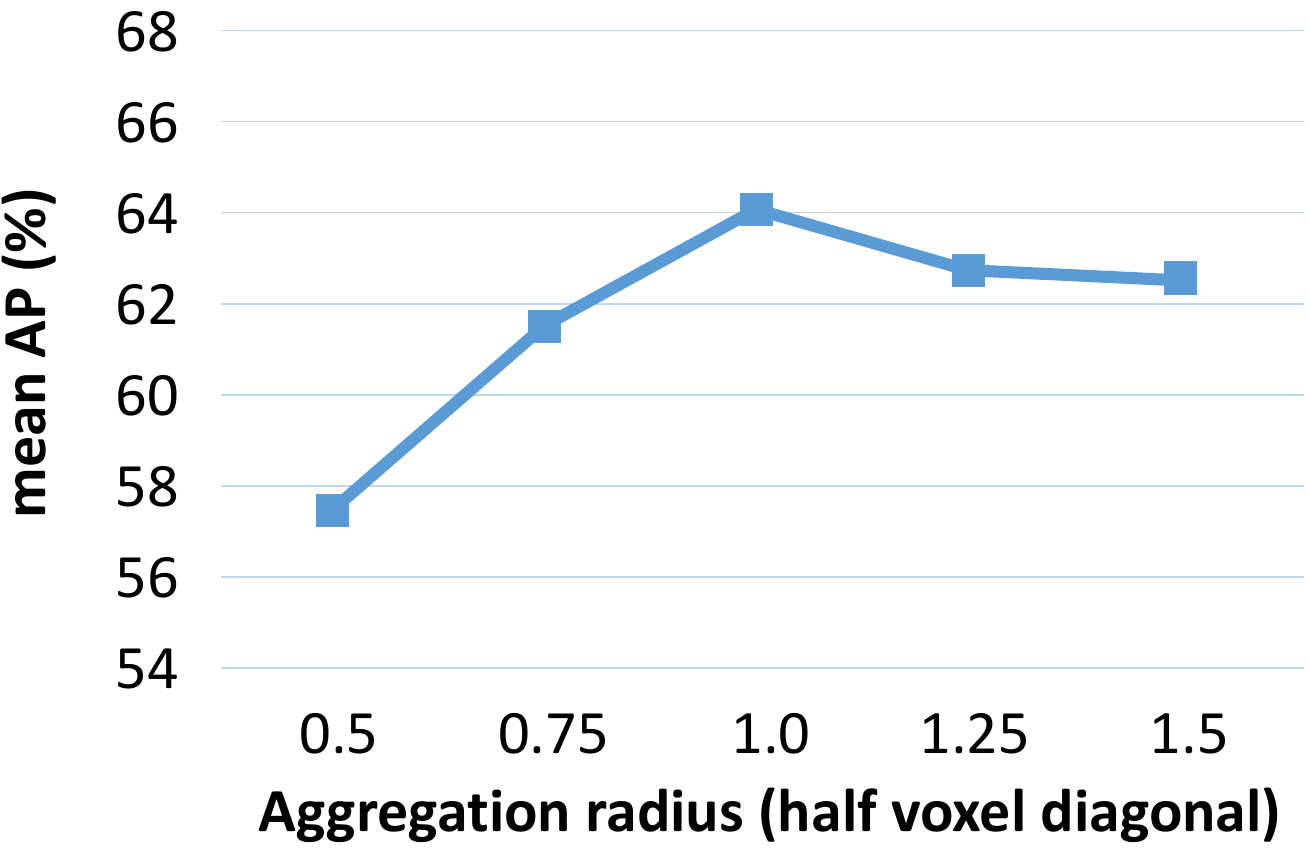}
\end{center}
\end{minipage}
\begin{minipage}[c]{0.50\linewidth}
\begin{center}
\setlength{\tabcolsep}{8.0pt}
\begin{tabular}{c|c}
    \toprule
    Aggregation methods & mAP \\
    \midrule
    Avg. pooling & 62.6 \\ 
    Nearest neighbor & 62.4 \\
    \midrule
    Interpolation & \textbf{64.1} \\
    \bottomrule
\end{tabular}
\end{center}
\end{minipage}
\caption{\textbf{Feature aggregation analysis.} \emph{Left:} mAP@0.25 on ScanNetV2 for varying aggregation radii (aggregation via interpolation). \emph{Right:} Comparisons of using different aggregation methods (radius = $1.0$)
}
\label{fig:aggre_radius}
\vspace{-0.1in}
\end{figure}

The radius $r$ in Eqn.~\ref{eqn:grid_kernel} controls how much features from neighboring points are aggregated to each voxel. 
We here investigate how it affects feature aggregation and in turn detection.
We set half the diagonal of a voxel cell as a unit of radius.
Fig.~\ref{fig:aggre_radius} (left) shows as the radius increases, the mAP reaches its peak at $1.0$ radius. 
However, embracing more points from a larger neighborhood, though introducing more context, could excessively reduces representative features from voxel's nearby points, thus blurs voxelized representation and hurts detection performance. 

We proceed with a second analysis on the influence of different feature aggregation methods.
We test two alternative aggregation methods, average pooling and nearest neighbor, which compute the activation on each voxel by averaging features of its neighboring points and by directly taking its nearest point's features, respectively (tested with $1.0$ radius). 
Fig.~\ref{fig:aggre_radius} (right) shows that aggregation via interpolation outperforms both alternatives. One possible explanation is that it not only accumulates appropriate context but also respects positional correspondence between voxels and points. 

\begin{figure}[b]
\begin{minipage}[c]{0.50\linewidth}
\begin{center}
    \includegraphics[width=1.03\linewidth]{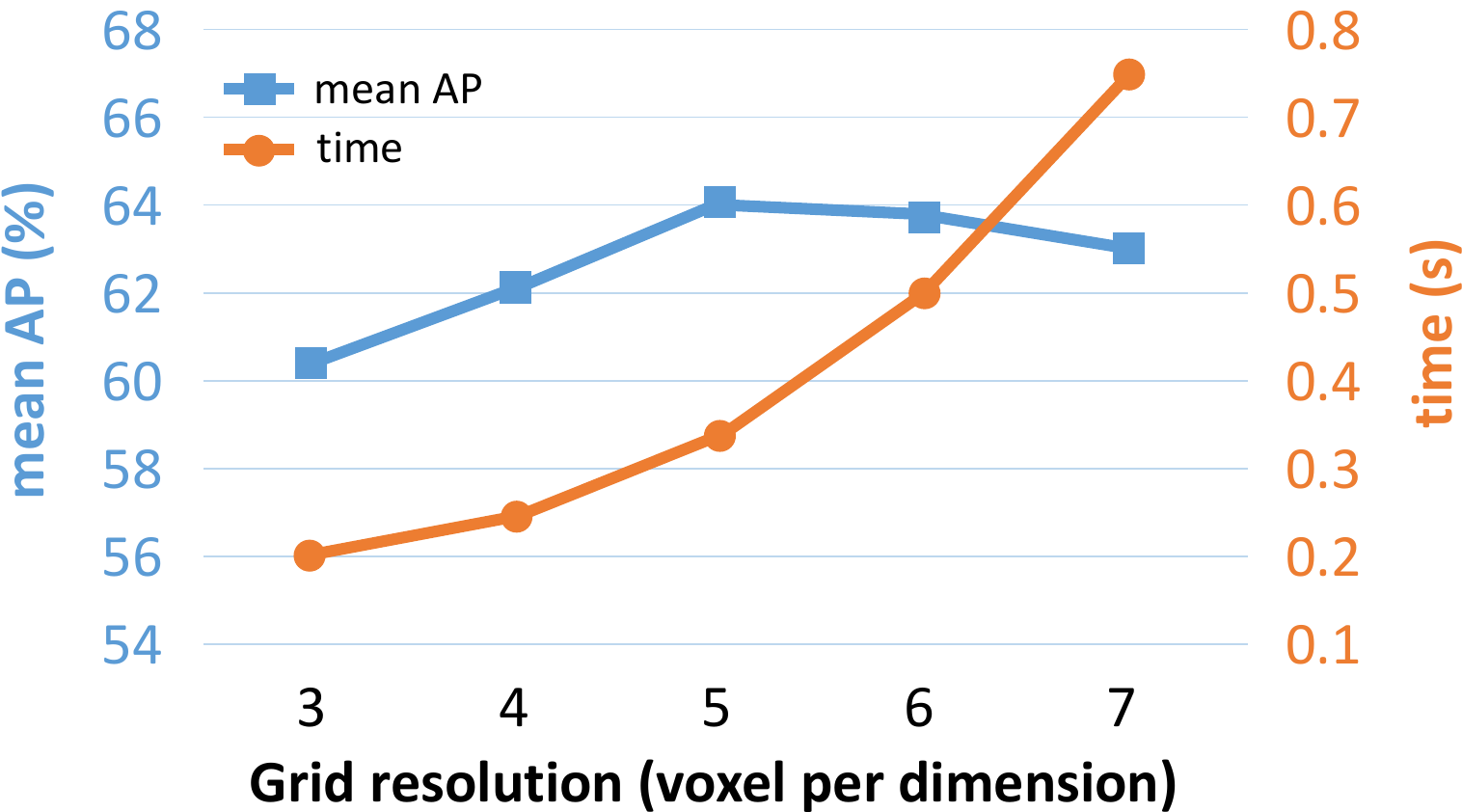}
\end{center}
\end{minipage}
\begin{minipage}[c]{0.50\linewidth}
\begin{center}
\setlength{\tabcolsep}{16.0pt}
\begin{tabular}{c|c}
	\toprule
	mini-CNNs & mAP \\
	\midrule
	1 \x\ (3\x3\x3) & 61.2 \\ \hline
	2 \x\ (1\x1\x1) & 58.1 \\ 	
	2 \x\ (3\x3\x3) & \textbf{64.1} \\ 
	2 \x\ (5\x5\x5) & 63.0 \\ 
	\bottomrule
\end{tabular}
\end{center}
\end{minipage}
\caption{\textbf{Analysis on grid resolution and mini-CNN architecture.} \emph{Left:} mAP@0.25 on ScanNetV2 for varying grid resolutions (convolve via two 3\x3\x3 convolutional layers). \emph{Right:} Comparisons of using different mini-CNN architectures (grid resolution = $5$)}
\label{fig:resolution_arch}
\end{figure}

\subsubsection{Effect of Grid Resolution}
\label{sec:exp_resolution}
The resolution of the rendered grid influences the quantization degree of local geometric structures and also computation overhead.
To determine its value, we compare detection performance and processing time (per scan) by using different grid resolutions. 
Fig.~\ref{fig:resolution_arch} (left) shows that
a proper resolution of 5\x5\x5 (voxels) gives the best detection performance. 
One explanation is that a lower resolution such as 3\x3\x3 introduces quantization artifacts during rendering, which obscures fine-grained geometry. 
In the contrary, projecting a point set with limited points to a grid of a higher resolution such as 7\x7\x7 would cause sparsity in voxel activation, which makes pattern learning difficult and increases processing time.
We thus adopt 5\x5\x5 as the grid resolution throughout our experiments.

\subsubsection{Effect of Mini-CNN Architecture}
We are interested in how different mini-CNN architectures influence local pattern abstraction.
Fig.~\ref{fig:resolution_arch} (right) provides detection performance of using different numbers and kernel sizes of convolutional layers.
One can see that using two 1\x1\x1 convolutional layers (somewhat like point-wise MLPs used by VoteNet) leads to a largely decreased mAP ($64.1$$\to$$58.1$), which is similar as the VoteNet performance ($58.6$).
This indicates that leveraging larger convolutional kernels to model spatial relations among voxels is the key to effectively capturing fine-grained geometric structures. 
In addition, two 3\x3\x3 convolutional layers (followed by a global max pooling) are sufficient for abstracting patterns in the grid. 
Since no noticeable improvement is observed by further increasing the number or kernel size of convolutional layers, we use the above settings throughout our experiment.

\vspace{-0.1in}
\subsubsection{Effect of Query Strategy}
\label{sec:exp_grouping}
We adopt cube query to construct local point sets for sensed regions. 
To validate our design choice, we further test with ball query as used by PointNet++~\cite{qi2017pointnet++}. 
For fair comparisons, we set the cube and ball to have the same volume in each SP module. 
Experiments show that either query strategy obtains satisfying results. 
Specifically, cube query performs slightly better than ball query: $64.1$ vs. $63.5$ in mAP.
We argue that cube query’s local neighborhood is more compatible with the rendered grid in shape compared to ball query's. 
This guarantees a more complete coverage of voxels when interpolating queried points onto the grid, which brings informative grid representation and thus facilitates pattern abstraction.
\vspace{-0.05in}
\begin{figure}
	\begin{center}
		\includegraphics[width=0.98\linewidth]{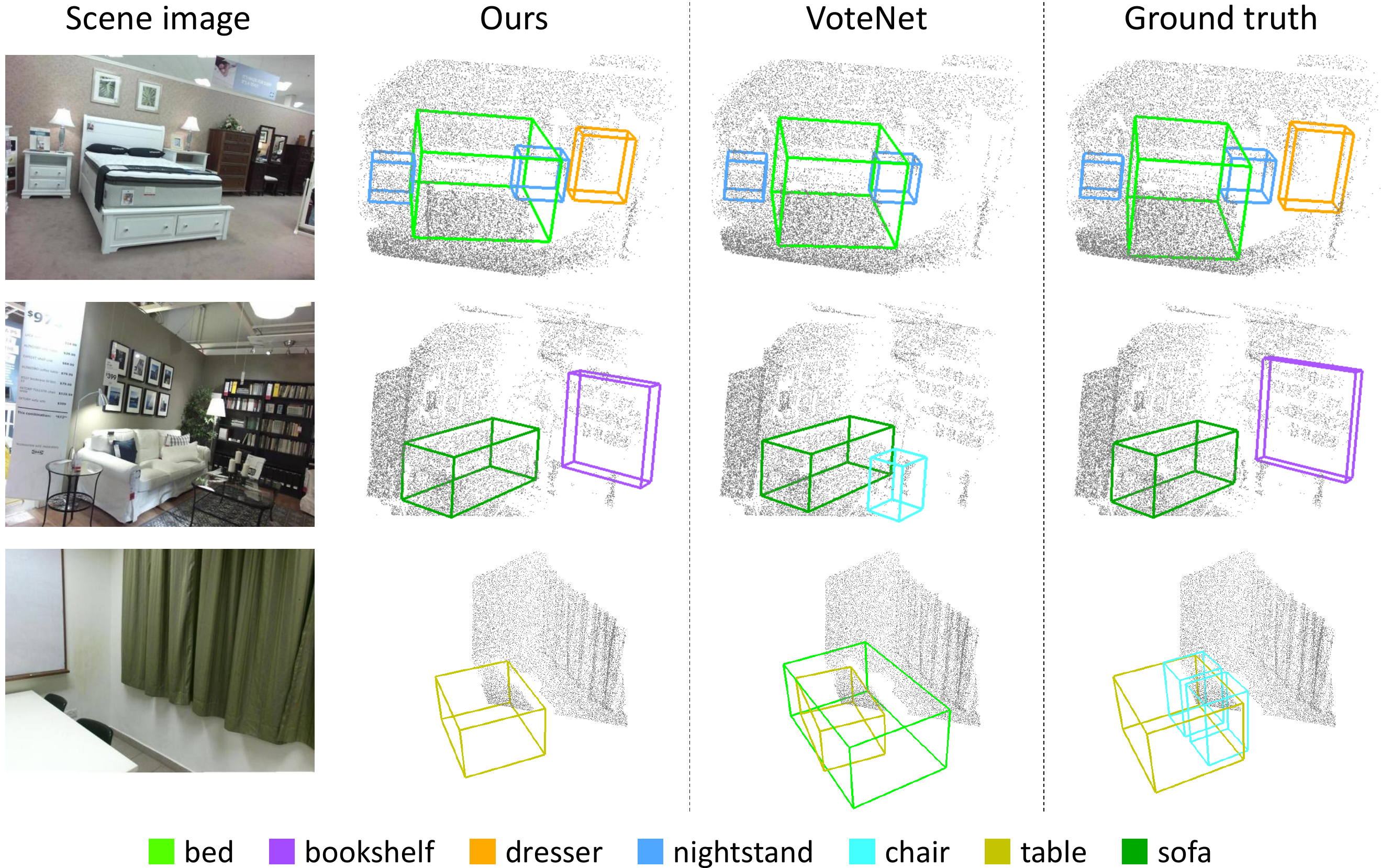}
	\end{center}
	\vspace{-0.2in}
	\caption{\textbf{Qualitative results of 3D object detection on SUN RGB-D}. Each row shows (from left to right): an image of the scene (not used by our network), 3D object detection by our model and by VoteNet, and ground-truth annotations}
	\label{fig:qualitative_sunrgbd}
\end{figure}

\begin{figure}
	\begin{center}
		\includegraphics[width=0.78\linewidth]{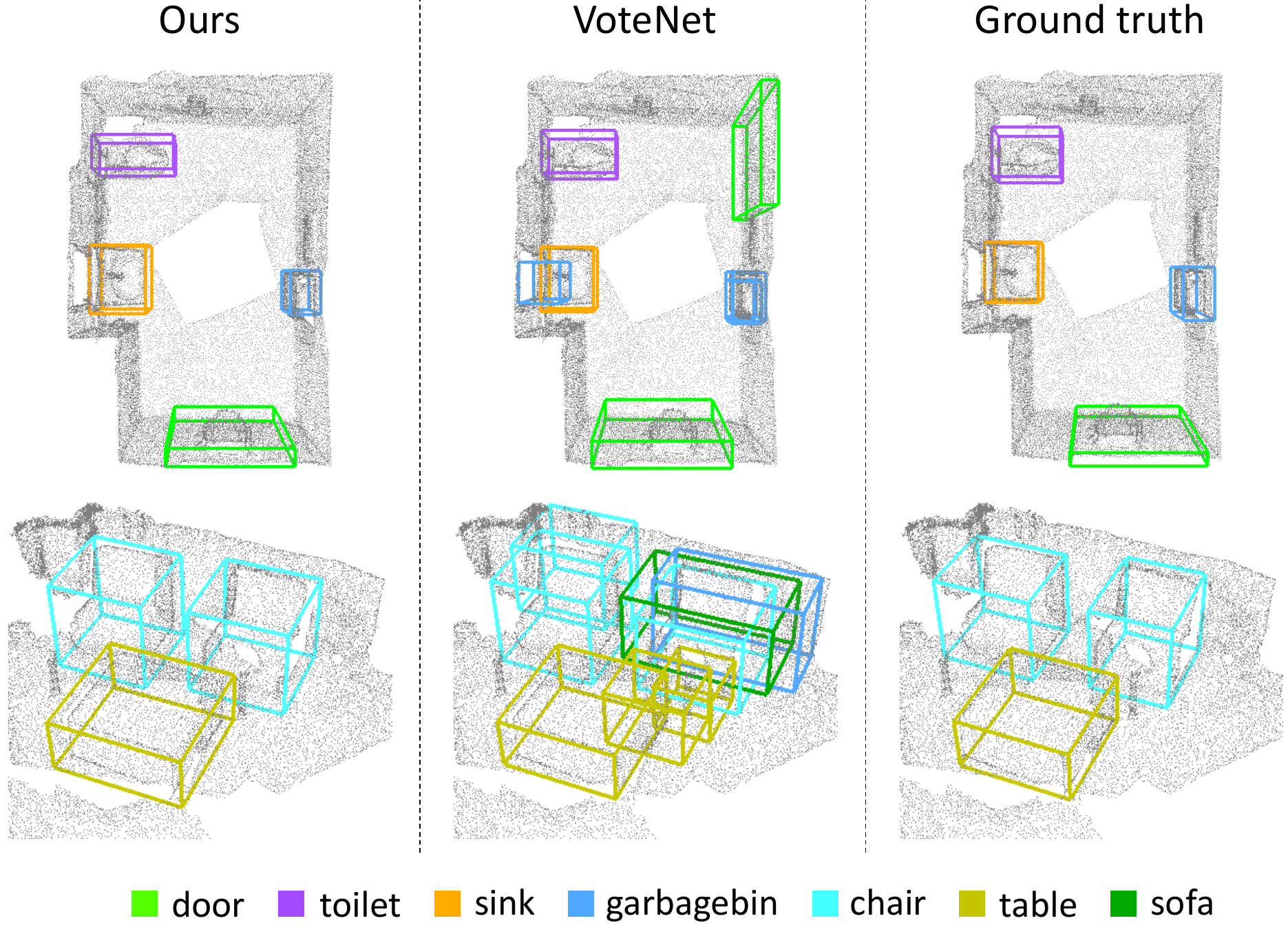}
	\end{center}
	\vspace{-0.2in}
	\caption{\textbf{Qualitative results of 3D object detection on ScanNetV2}. Each row shows (from left to right): 3D object detection by our model and by VoteNet, and ground-truth annotations}
	\label{fig:qualitative_scannet}
	\vspace{-0.35in}
\end{figure}

\subsection{Qualitative Results and Discussion}
In Fig.~\ref{fig:qualitative_sunrgbd}, we provide qualitative results of our model and VoteNet on SUN RGB-D to show how extracting point features through LGR-Net benefits detection in various ways.
The first two examples show that a clustered dresser/bookshelf is missed by VoteNet, while our model is able to collect sufficient geometric cues at corresponding positions, thus boosts confidence to recognize the object.
The second example also reveals the strengths of our model in avoiding false positives like the chair as that in VoteNet output, indicating our model is superior in learning informative region features to resolve local ambiguities. 
The above results evidence that our model effectively promotes extracted features from point clouds, and thus helps localize and recognize objects.
The last example shows a less successful prediction where an extremely partial observation of chairs is given.
Similar advantages of our approach are also revealed by qualitative results on ScanNetV2, as shown in Fig.~\ref{fig:qualitative_scannet}.

\section{Conclusions}
\vspace{-0.05in}
In this work, we propose a novel Local Grid Rendering (LGR) operation which allows using small-size CNNs to effectively abstract fine-grained geometric structures while preserving computational efficiency. 
A simple yet efficient backbone LGR-Net for point cloud feature extraction is further introduced based on the LGR operation. 
We apply the LGR-Net to object detection.
With only point input, our model shows significant improvements over prior arts.
Our proposed LGR-Net is generic.
In future work we intend to utilize it in downstream applications such as 3D instance segmentation in point clouds. 

\clearpage
%
%
\bibliographystyle{splncs04}
\bibliography{egbib}

\begin{thebibliography}{10}
\providecommand{\url}[1]{\texttt{#1}}
\providecommand{\urlprefix}{URL }
\providecommand{\doi}[1]{https://doi.org/#1}

\bibitem{avetisyan2019scan2cad}
Avetisyan, A., Dahnert, M., Dai, A., Savva, M., Chang, A.X., Nie{\ss}ner, M.:
  Scan2cad: Learning cad model alignment in rgb-d scans. In: CVPR. pp.
  2614--2623 (2019)

\bibitem{bronstein2010scale}
Bronstein, M.M., Kokkinos, I.: Scale-invariant heat kernel signatures for
  non-rigid shape recognition. In: CVPR. pp. 1704--1711 (2010)

\bibitem{chen2017multi}
Chen, X., Ma, H., Wan, J., Li, B., Xia, T.: Multi-view 3d object detection
  network for autonomous driving. In: CVPR. pp. 1907--1915 (2017)

\bibitem{choy20194d}
Choy, C., Gwak, J., Savarese, S.: 4d spatio-temporal convnets: Minkowski
  convolutional neural networks. In: CVPR. pp. 3075--3084 (2019)

\bibitem{dai2017scannet}
Dai, A., Chang, A.X., Savva, M., Halber, M., Funkhouser, T., Nie{\ss}ner, M.:
  Scannet: Richly-annotated 3d reconstructions of indoor scenes. In: CVPR. pp.
  5828--5839 (2017)

\bibitem{engelcke2017vote3deep}
Engelcke, M., Rao, D., Wang, D.Z., Tong, C.H., Posner, I.: Vote3deep: Fast
  object detection in 3d point clouds using efficient convolutional neural
  networks. In: ICRA. pp. 1355--1361 (2017)

\bibitem{golovinskiy2009shape}
Golovinskiy, A., Kim, V.G., Funkhouser, T.: Shape-based recognition of 3d point
  clouds in urban environments. In: ICCV. pp. 2154--2161 (2009)

\bibitem{gomes2013efficient}
Gomes, R.B., da~Silva, B.M.F., de~Medeiros~Rocha, L.K., Aroca, R.V., Velho,
  L.C.P.R., Gon{\c{c}}alves, L.M.G.: Efficient 3d object recognition using
  foveated point clouds. Computers \& Graphics  \textbf{37}(5),  496--508
  (2013)

\bibitem{gomez2016pl}
Gomez-Ojeda, R., Briales, J., Gonzalez-Jimenez, J.: Pl-svo: Semi-direct
  monocular visual odometry by combining points and line segments. In: IROS.
  pp. 4211--4216 (2016)

\bibitem{graham20183d}
Graham, B., Engelcke, M., van~der Maaten, L.: 3d semantic segmentation with
  submanifold sparse convolutional networks. In: CVPR. pp. 9224--9232 (2018)

\bibitem{he2017mask}
He, K., Gkioxari, G., Doll{\'a}r, P., Girshick, R.: Mask r-cnn. In: ICCV. pp.
  2961--2969 (2017)

\bibitem{hou20193d}
Hou, J., Dai, A., Nie{\ss}ner, M.: 3d-sis: 3d semantic instance segmentation of
  rgb-d scans. In: CVPR. pp. 4421--4430 (2019)

\bibitem{klokov2017escape}
Klokov, R., Lempitsky, V.: Escape from cells: Deep kd-networks for the
  recognition of 3d point cloud models. In: ICCV. pp. 863--872 (2017)

\bibitem{lahoud20172d}
Lahoud, J., Ghanem, B.: 2d-driven 3d object detection in rgb-d images. In:
  CVPR. pp. 4622--4630 (2017)

\bibitem{lan2019modeling}
Lan, S., Yu, R., Yu, G., Davis, L.S.: Modeling local geometric structure of 3d
  point clouds using geo-cnn. In: CVPR. pp. 998--1008 (2019)

\bibitem{lang2019pointpillars}
Lang, A.H., Vora, S., Caesar, H., Zhou, L., Yang, J., Beijbom, O.:
  Pointpillars: Fast encoders for object detection from point clouds. In: CVPR.
  pp. 12697--12705 (2019)

\bibitem{le2018pointgrid}
Le, T., Duan, Y.: Pointgrid: A deep network for 3d shape understanding. In:
  CVPR. pp. 9204--9214 (2018)

\bibitem{litany2015asist}
Litany, O., Remez, T., Freedman, D., Shapira, L., Bronstein, A., Gal, R.:
  Asist: automatic semantically invariant scene transformation. Computer Vision
  and Image Understanding  \textbf{157},  284--299 (2017)

\bibitem{mao2019interpolated}
Mao, J., Wang, X., Li, H.: Interpolated convolutional networks for 3d point
  cloud understanding. In: ICCV. pp. 1578--1587 (2019)

\bibitem{Indoor2012}
Nan, L., Xie, K., Sharf, A.: A search-classify approach for cluttered indoor
  scene understanding. ACM TOG  \textbf{31}(6),  1--10 (2012)

\bibitem{oh2002development}
Oh, Y.J., Watanabe, Y.: Development of small robot for home floor cleaning. In:
  SICE. vol.~5, pp. 3222--3223 (2002)

\bibitem{park2008multiple}
Park, Y., Lepetit, V., Woo, W.: Multiple 3d object tracking for augmented
  reality. In: ISMAR. pp. 117--120 (2008)

\bibitem{qi2019deep}
Qi, C.R., Litany, O., He, K., Guibas, L.J.: Deep hough voting for 3d object
  detection in point clouds. In: ICCV. pp. 9277--9286 (2019)

\bibitem{qi2018frustum}
Qi, C.R., Liu, W., Wu, C., Su, H., Guibas, L.J.: Frustum pointnets for 3d
  object detection from rgb-d data. In: CVPR. pp. 918--927 (2018)

\bibitem{qi2017pointnet}
Qi, C.R., Su, H., Mo, K., Guibas, L.J.: Pointnet: Deep learning on point sets
  for 3d classification and segmentation. In: CVPR. pp. 652--660 (2017)

\bibitem{qi2016volumetric}
Qi, C.R., Su, H., Nie{\ss}ner, M., Dai, A., Yan, M., Guibas, L.J.: Volumetric
  and multi-view cnns for object classification on 3d data. In: CVPR. pp.
  5648--5656 (2016)

\bibitem{qi2017pointnet++}
Qi, C.R., Yi, L., Su, H., Guibas, L.J.: Pointnet++: Deep hierarchical feature
  learning on point sets in a metric space. In: NeurIPS. pp. 5099--5108 (2017)

\bibitem{ren2016three}
Ren, Z., Sudderth, E.B.: Three-dimensional object detection and layout
  prediction using clouds of oriented gradients. In: CVPR. pp. 1525--1533
  (2016)

\bibitem{riegler2017octnet}
Riegler, G., Osman~Ulusoy, A., Geiger, A.: Octnet: Learning deep 3d
  representations at high resolutions. In: CVPR. pp. 3577--3586 (2017)

\bibitem{shi2019pointrcnn}
Shi, S., Wang, X., Li, H.: Pointrcnn: 3d object proposal generation and
  detection from point cloud. In: CVPR. pp. 770--779 (2019)

\bibitem{song2015sun}
Song, S., Lichtenberg, S.P., Xiao, J.: Sun rgb-d: A rgb-d scene understanding
  benchmark suite. In: CVPR. pp. 567--576 (2015)

\bibitem{song2014sliding}
Song, S., Xiao, J.: Sliding shapes for 3d object detection in depth images. In:
  ECCV. pp. 634--651 (2014)

\bibitem{song2016deep}
Song, S., Xiao, J.: Deep sliding shapes for amodal 3d object detection in rgb-d
  images. In: CVPR. pp. 808--816 (2016)

\bibitem{wang2019dynamic}
Wang, Y., Sun, Y., Liu, Z., Sarma, S.E., Bronstein, M.M., Solomon, J.M.:
  Dynamic graph cnn for learning on point clouds. ACM TOG  \textbf{38}(5),
  1--12 (2019)

\bibitem{wu20153d}
Wu, Z., Song, S., Khosla, A., Yu, F., Zhang, L., Tang, X., Xiao, J.: 3d
  shapenets: A deep representation for volumetric shapes. In: CVPR. pp.
  1912--1920 (2015)

\bibitem{xu2018pointfusion}
Xu, D., Anguelov, D., Jain, A.: Pointfusion: Deep sensor fusion for 3d bounding
  box estimation. In: CVPR. pp. 244--253 (2018)

\bibitem{xu2019grid}
Xu, Q.: Grid-gcn for fast and scalable point cloud learning. arXiv preprint
  arXiv:1912.02984  (2019)

\bibitem{yi2019gspn}
Yi, L., Zhao, W., Wang, H., Sung, M., Guibas, L.J.: Gspn: Generative shape
  proposal network for 3d instance segmentation in point cloud. In: CVPR. pp.
  3947--3956 (2019)

\bibitem{zhou2018voxelnet}
Zhou, Y., Tuzel, O.: Voxelnet: End-to-end learning for point cloud based 3d
  object detection. In: CVPR. pp. 4490--4499 (2018)

\end{thebibliography}

\end{document}